\documentclass[sigconf]{acmart}

\usepackage{algorithmic}
\usepackage{algorithm}
\usepackage{multirow}

\AtBeginDocument{%
  \providecommand\BibTeX{{%
    \normalfont B\kern-0.5em{\scshape i\kern-0.25em b}\kern-0.8em\TeX}}}

\setcopyright{acmcopyright}
\copyrightyear{2022}
\acmYear{2022}
\acmDOI{10.1145/1122445.xxxx}

\acmConference[ACM '22]{ACM '22: The xx ACM International Conference on XXX}{XX. xx--xx, 2022}{USA}
\acmBooktitle{ACM '22: The xx ACM International Conference on XXX, XX. xx--xx, 2022, USA}
\acmPrice{15.00}
\acmISBN{978-1-4503-XXXX-X/18/06}

\begin{document}

\title{Robust Time Series Dissimilarity Measure for Outlier Detection and Periodicity Detection}


\author{Xiaomin Song, Qingsong Wen, Yan Li, Liang Sun}
\affiliation{%
  \institution{DAMO Academy, Alibaba Group,
Bellevue, USA}
  \city{}
  \state{}
  \country{}
}







\renewcommand{\shortauthors}{Song and Wen, et al.}

\begin{abstract}
Dynamic time warping (DTW) is an effective dissimilarity measure in many time series applications. Despite its popularity, it is prone to noises and outliers, which leads to singularity problem and bias in the measurement. The time complexity of DTW is quadratic to the length of time series, making it inapplicable in real-time applications. In this paper, we propose a novel time series dissimilarity measure named RobustDTW to reduce the effects of noises and outliers. Specifically, the RobustDTW estimates the trend and optimizes the time warp in an alternating manner by utilizing our designed temporal graph trend filtering. To improve efficiency, we propose a multi-level framework that estimates the trend and the warp function at a lower resolution, and then repeatedly refines them at a higher resolution. Based on the proposed RobustDTW, we further extend it to periodicity detection and outlier time series detection. Experiments on real-world datasets demonstrate the superior performance of RobustDTW compared to DTW variants in both outlier time series detection and periodicity detection.
\end{abstract}




\maketitle

\section{Introduction}

Nowadays, time series (TS) signal processing and mining have received lots of research interests~\cite{huang2019dsanet,esling2012time,angelosante2011sparse,zhang2021cloudrca,wen2020fastrobustSTL}. Among them, 
dynamic time warping (DTW)~\cite{SakoeChiba71,yuan2019locally,paparrizos2020debunking,zhang2012fast,pratzlich2016memory} is a popular method to compute the dissimilarity between two TS by finding the optimal alignment. It has been widely employed in many tasks involving distance computation such as TS similarity search~\cite{search:dtw:kdd:2012}, outlier detection~\cite{DTW:outlier:2019:Diab,benkabou2018unsupervised}, clustering and classification~\cite{WDTW:classification:PR:2011}, periodicity detection~\cite{elfeky2005warp}, etc. Due to its simplicity and versatility, it has been an effective tool in many areas, including speech recognition, signal processing, machine learning, bioinformatics, etc.




Although DTW solves the time warping problem, it suffers from some limitations. Firstly, in practice time series data is often contaminated by noises and outliers. The DTW only considers the stretching or shrinking along the time axis but ignores to handle noises and outliers in the value axis, which may bias the distance between time series and lead to the singularity problem where a single point in one TS is mapped to a large subsection in the other TS. The robustness of time series distance is crucial in many time series applications, such as outlier time series detection and periodicity detection. For example, in outlier TS detection the goal is to identity outlier TS different from others. In this scenario, some sporadic noises or outliers in a normal time series may significantly increase the distance between it and others using DTW, and thus reporting a false positive outlier TS. 
Secondly, the time and space complexity of DTW are quadratic to the length of TS, which makes it difficult to be applied in long sequence analysis.
Recently, some works ~\cite{Boyd:DTW:2020,LI2020:adaptive:dtw,salvador2007toward,junior2007improved,han2018accurate} are proposed to deal with either singularity or complexity problem.
However, existing methods cannot robustly and efficiently address both challenges simultaneously.





In this paper, we propose a novel dissimilarity measure called RobustDTW, which addresses the aforementioned challenges. To handle the noises and outliers in TS, we propose a general temporal graph trend filtering to estimate the true signal. Compared with classical DTW and its variants, in RobustDTW we optimize the time warp function and perform the trend filtering simultaneously. To further accelerate computation, we propose a multi-level framework to learn the time warp function and perform trend filtering recursively. Specifically, we perform downsampling recursively to get different levels of representations of TS. Then we start from a lower resolution representation to calculate the time warp function. And the noises and outliers are handled properly by the proposed temporal graph trend filtering, where the graph is constructed based on the learned time warp function. 
Meanwhile, the singularity problem is mitigated in this multi-level framework. To the best of our knowledge, this is the first paper to design a general temporal graph trend filtering extended for time warping, which is further incorporated into a novel multi-level framework for efficient computation. 

To demonstrate the effectiveness of RobustDTW, we apply it in outlier time series detection by identifying abnormal TS given a set of TS. By integrating the proposed RobustDTW with the popular local outlier factor (LOF) algorithm~\cite{breunig2000lof,wang2011efficient,liu2020data}, we can successfully identify these abnormal TS from noisy data.
To further justify our proposed RobustDTW algorithm, we extend it to the periodicity detection task. 
Empirically it outperforms state-of-the-art periodicity detection algorithms~\cite{WenRobustPeriod20,vlachos2005periodicity,Mitsa2010}, as it can robustly handle non-stationary TS with noises, outliers, and complicated periodic patterns.

\section{Related Work}

One issue of DTW is the singularity~\cite{Boyd:DTW:2020} introduced by noises and outliers in time series, which causes many time points of one TS to be erroneously mapped onto a single point of the other TS. To mitigate this problem, the derivative DTW~\cite{keogh2001derivative} is proposed to compute the shape information by considering the first derivative of the sequences. Recently, \cite{Boyd:DTW:2020} proposes to compute the optimal time warp function by solving an optimization problem which considers the misalignment of time warping with two penalization terms for the cumulative warping and the instantaneous rate of time warping. 
In \cite{LI2020:adaptive:dtw}, an adaptive constrained DTW is proposed by introducing two adaptive penalties to mitigate the singularity problem.

Another issue of classical DTW algorithm is the quadratic time and space complexity which makes it a bottleneck in many applications. 
To mitigate this problem, the most widely used approach is FastDTW~\cite{salvador2007toward}, which is an efficient approximate algorithm with linear time and space complexity. It adopts a multi-level approach to compute the optimal time warp function, from the coarsest level with repeated refining.
Similar works considering applying wavelet transform in the multi-level framework can be found in~\cite{wavelet:DTW,junior2007improved,han2018accurate}. 
Another way to speed up DTW is to use searching constraints.
In~\cite{keogh2005exact}, the small constraint and lower bounding are utilized to skip some computation in DTW.
Some empirical comparisons are reported in~\cite{wu2020fastdtw}. Generally, when the length of time series is relatively small (less than several thousand), and the warping path is close to the diagonal path, the constrain-based methods outperforms. However, as the time series length increases and the warping path deviate significantly, the multi-level methods are better choices~\cite{salvador2007toward}. 

Besides, some recent works extend DTW to differentiable versions like Soft-DTW~\cite{cuturi2017soft} and DP-DTW~\cite{chang2021learning}, which can be integrated into deep learning networks. Note that we are not dealing with differentiable DTW in this paper. Instead, also different from excising work focusing on either mitigating the singularity problem or reducing time and space complexity, we aim to solve both problems for the common DTW algorithm in one-shot by proposing a novel RoubustDTW algorithm, which is further illustrated in periodicity detection and outlier time series detection applications to demonstrate the superior
performance of the proposed algorithm.

\vspace{-2mm}
\section{Methodology}\label{section:method}
\vspace{-1mm}

\subsection{Proposed RobustDTW Algorithm} \label{subsection:robustdtw} 

To make DTW robust to noise and outliers while preserving its flexibility of dynamic index matching, we employ the robust trend to filter out the noises and outliers. Specifically, we propose to estimate the time warp function and detrend TS in an alternating manner. We iteratively conduct the following operations: (1) fix the time warp function and estimate the detrended TS $u$ and $v$; (2) fix the estimated detrended TS $u$ and $v$, and estimate the time warp function $\varphi(t)$ based on $u$ and $v$. 

To speed up the computation of RobustDTW, we utilize the similarity of the shapes and alignments of TS pairs among different resolutions, and combine alternating of DTW align adjustment and trend estimate with the progress from low resolution to high resolution. 
During this multi-level progress, low resolution trend results are the starting values for high resolution estimation, as low resolution index alignment can suggest the constraint of path searching space for high resolutions. 
We elaborate the detailed procedure in the following six steps. In the following, we denote the two TS to be compared as $\mathbf{x}$ and $\mathbf{y}$.

\vspace{1mm}
{\hspace{-3.5mm}\bf Step 1: Robust Self-Detrending}\\
To roughly remove the effects of outliers and noises, we first adopt robust trend filtering~\cite{wen2019robusttrend} on individual TS. Formally, we denote the TS of length $n$ as $\mathbf{y}=[y_1, y_2, \cdots, y_{n}]^T$, which can be decomposed into trend and residual components as follows: 
\begin{equation}\label{eq:Univariate} 
y_t = v_t + r_t \quad\text{or}\quad \mathbf{y} = \mathbf{v} + \mathbf{r},
\end{equation}
where $\mathbf{v}=[v_1, v_2, \cdots, v_{n}]^T$ denotes the trend component, and $\mathbf{r} = [r_1, r_2, \cdots, r_{n}]^T$ denotes the residual component, which also contains noises and outliers. Generally, in trending filtering we want the trend to be smooth and the residual to be small. 

We notice that there are various trend filtering algorithms proposed in the literature~\cite{alexandrov2012review}, such as the Hodrick–Prescott (H-P) filter~\cite{hodrick1997postwar}, the $\ell_1$ trend filtering~\cite{kim2009ell_1}, median filtering~\cite{Wen:median:filter}, etc. In this paper, we adopt the RobustTrend filter~\cite{wen2019robusttrend}, which is robust to outliers and can capture both slow and abrupt trend changes. In the following, we use $\mathbf{u}$ and $\mathbf{v}$ to denote the trend extracted from the original time series $\mathbf{x}$ and $\mathbf{y}$, respectively.

\vspace{1mm}
{\hspace{-3.5mm}\bf Step 2: Multi-Level Representation}\\
Multi-level representations is achieved by downsampling filtered trend TS from Step 1 by factor of 2 iteratively as follows
%
\begin{equation}\label{eq:MultiLevel} 
\mathbf{u}_{\ell-1} \xrightarrow[]{downsample} \mathbf{u}_{\ell-1}^\downarrow = \mathbf{u}_{\ell},~\mathbf{v}_{\ell-1} \xrightarrow[]{downsample} \mathbf{v}_{\ell-1}^\downarrow = \mathbf{v}_{\ell}
\end{equation}
where $\mathbf{u}_{1}=\mathbf{u}, \mathbf{v}_{1}=\mathbf{v}, \ell=2,...,t$. 
Specifically, the level 1 representation is the original filtered trend, and downsample level 1 by factor of 2 to generate the level 2 representation. 
This procedure is conducted until the level $t$ representation is obtained, where $t$ is the total levels in the framework. 
Note that higher level representations correspond to lower resolution TS.

{\hspace{-3.5mm}\bf Step 3: Projection and Upsampling}\\
Starting from level $t$ TS, call FastDTW~\cite{salvador2007toward} to get warping index alignment $\boldsymbol{\pi}_{t}$ and use current level representations as the base trend estimation. In the subsequent iterations, upsample previous level's warping path ($\boldsymbol{\pi}_{\ell}$) by 2 and add extra searching width defined by parameter radius $r$ to generate the searching constraint, which is called projection. That is
\begin{equation}\label{eq:project_upsmaple} \notag
\boldsymbol{\pi}_{\ell} \xrightarrow[]{upsample} \boldsymbol{\pi}_{\ell}^\uparrow, ~ \mathcal{A}(\mathbf{u},\mathbf{v})_{\ell-1}=\{\text{path set}|\text{center}~ \boldsymbol{\pi}_{\ell}^\uparrow, \text{ radius}~ r \}.
\end{equation}


\vspace{1mm}
{\hspace{-3.5mm}\bf Step 4: Time Warping Alignment}\\
Refine the warping path by running DTW with projected warping constrain ($\mathcal{A}(\mathbf{u},\mathbf{v})_{\ell-1}$) to obtain the time warp function ($\varphi(t)$) and path ($\boldsymbol{\pi}_{\ell-1}$) at current level as 
\begin{equation}\label{eq:twa} 
\boldsymbol{\pi}_{\ell-1}, \varphi(t) = \min_{\boldsymbol{\pi} \in \mathcal{A}(\mathbf{u},\mathbf{v})_{\ell-1}} \sqrt{\sum_{(t,\varphi(t))\in\boldsymbol{\pi}}d(u_t,v_{\varphi(t)})^2  }.
\end{equation}






\vspace{1mm}
{\hspace{-3.5mm}\bf Step 5: Temporal Graph Detrending}\\
In this step, we aim to estimate the detrended TS $\mathbf{u}$ and $\mathbf{v}$ more accurately by considering not only its neighbors within itself but also its similar peer time series. 
Let $G = (V, E)$ be an graph, with vertices $V = \{1, \cdots, n\}$ and undirected edges $E = \{e_1, \cdots, e_s\}$, and suppose that we observe $\mathbf{y} = [y_1,\cdots, y_n]^T \in R^n$ over each node. Similarly, the $k$th order graph trend filtering estimates $\boldsymbol{\tau}=[\tau_1, \cdots, \tau_n]^T$ can be obtained by solving: 
\begin{equation}\label{eq:GTF} 
\text{argmin}_{\tau\in R^n} \ \frac{1}{2} \|\mathbf{y}-\boldsymbol{\tau}\|_2^2 + \lambda \| \Delta^{(k+1)} \boldsymbol{\tau} \|_1,
\end{equation}
where $\Delta^{(k+1)}$ is the graph difference operator of order $k+1$. 
When $k=0$, the 1st order graph difference operator $\Delta^{(1)}$ penalizes all local differences on all edges as
$
\|\Delta^{(1)} \boldsymbol{\tau} \|_1 = \sum_{(i,j)\in E} | \tau_i - \tau_j |. $
We can represent $\Delta^{(1)}$ in a matrix form as $\Delta^{(1)}\in \{-1,0,1\}^{s\times n}$ where $s=|E|$, i.e., number of edges. Specifically, let $e_{\ell}=(i,j)$, then $\Delta^{(1)}$ has the $\ell$th row as
$\Delta^{(1)}_{\ell} = [0, \cdots, -1, \cdots, 1, \cdots, 0]$,
i.e., $\Delta^{(1)}_{\ell}$ has -1 at the $i$th position, and 1 at the $j$th position. Similar to trend filtering on univariate TS, we can define the higher order graph difference operators recursively. In particular, the graph difference operators defined above reduce to the ones defined on the univariate TS in which $V=\{1,2,\cdots,n\}$ and $E=\{(i,i+1): i=1,2,\cdots,n-1\}$. 

\begin{figure}[t!]
    \centering
    \includegraphics[width=0.6\linewidth]{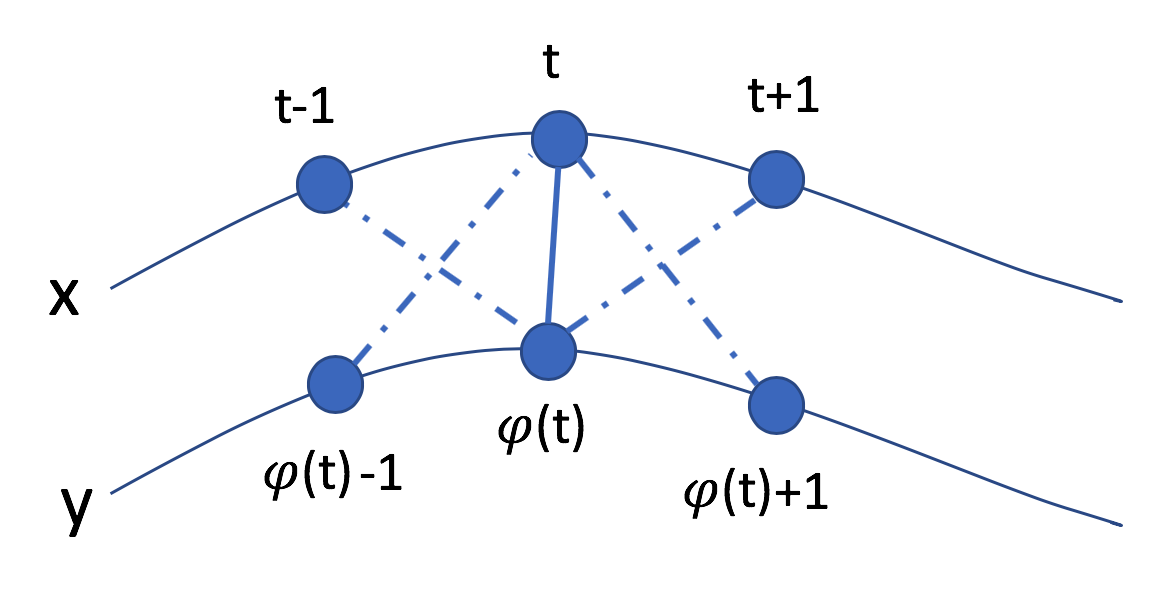}
    \vspace{-4mm}
    \caption{\small Illustration of the graph construction between two TS of the RobustDTW. Each data points are connected not only to their neighbors in the same TS but also to their peers of the other TS which are aligned by DTW. The time stamps on \textbf{x} are $t-1$, $t$, $t+1$, and $\varphi$(t) is the mapping of index from \textbf{x} to \textbf{y}. So the aligned points are $t-1$, $t$, $t+1$ on \textbf{x} to $\varphi(t)-1$, $\varphi(t)$, $\varphi(t)+1$ on \textbf{y}.}
    \label{fig:3-graph-detrending}
    \vspace{-4mm}
\end{figure}

In this paper, we design a general temporal graph detrending for multivariate TS by extending the idea of the graph-based detrending in Eq.~\eqref{eq:GTF}. 
The key idea is to incorporate the relationship between TS in detrending which can also deal with the lagging effect adaptively. Another distinguishing feature is that we introduce weight for each edge in the graph, instead of only allowing binary weight as in \cite{wang2016trend}. 
Specifically, we construct the graph of two TS based on the alignment at step 4. In the constructed graph $G$, we have $m+n$ vertices, and each vertex corresponds to a time point in $\mathbf{x}\in R^m$ and $\mathbf{y}\in R^n$. For notation simplicity, we denote the set of vertices as $E=\{x_1, x_2, \cdots, x_m, y_1, y_2, \cdots, y_n\}$. Next, we describe how to construct edges and their corresponding weights. Firstly, each vertex $x_t$ is connected its left neighbor $x_{t-1}$ and right neighbor $x_{t+1}$. Secondly, $x_t$ should be connected to its peer in $\mathbf{y}$, i.e., $y_{\varphi(t)}$. Thirdly, to avoid errors introduced in DTW of step 4, we construct more edges to improve robustness. For the edge $x_t \leftrightarrow y_{\varphi(t)}$, we introduce additional $4$ edges, including $x_t \leftrightarrow  y_{\varphi(t)-1}$, $x_t \leftrightarrow  y_{\varphi(t)+1}$, $x_{t-1} \leftrightarrow  y_{\varphi(t)}$, $x_{t+1} \leftrightarrow  y_{\varphi(t)}$, as illustrated in the Fig.~\ref{fig:3-graph-detrending}. Note that in Fig.~\ref{fig:3-graph-detrending} we only consider the direct one neighbor of $x_t$ and $y_{\varphi(t)}$. In practice, we can increase the size of the neighborhood based on the noises and outliers of data accordingly.

After constructing the graph $G$ for a pair of TS, the designed temporal graph detrending is computed as:
\vspace{-2mm}
\begin{equation}\label{eq:graph_detrend}
\mathbf{u, \!v}\!\!=\!\text{arg}\!\min\limits_{\mathbf{u, v}}\!\frac{1}{2} \|\mathbf{w}\!- [\mathbf{x}; \mathbf{y}]\|_2^2 \!+\! \lambda_1^{\textrm{GD}} \| D^{(1)}_\textrm{G} \!\mathbf{w}\|_1\!+\! \lambda_2^{\textrm{GD}} \| D^{(2)}_\textrm{G} \!\mathbf{w}\|_1\!,
\vspace{-2mm}
\end{equation}
where $\textbf{w}=[\textbf{u}; \textbf{v}]$ is the concatenate of input $\textbf{u}$ and $\textbf{v}$, 
${D}^{(1)}_\textrm{G}$ and ${D}^{(2)}_\textrm{G}$ are the 1st and 2nd graph difference operators 
used to capture abrupt and slow trend changes. Note that both self-detrending \eqref{eq:pre_detrend} and temporal graph detrending \eqref{eq:graph_detrend} can be efficiently solved by the alternating direction of multipliers (ADMM) algorithm~\cite{boyd2011distributed}.

\vspace{1mm}
{\hspace{-3.5mm}\bf Step 6: Iterative Processing}\\
To achieve better performance, we repeat Steps 3 to 5 to update the time warp function and the trend estimates until convergence. 
Furthermore, when we compute the time warp function in the higher resolution, we use the time warp function computed in the previous step as the initialization, which significantly reduces computational complexity. 

\section{Experiments and Applications} \label{sec:application-experiment}



\subsection{Setting and Efficiency Comparison} 

Although RobustDTW involves several parameters, we observe that empirically parameter tuning is not complex, as summarized below. For the temporal graph detrending, larger $\lambda_1$ and $\lambda_2$ would yield a smoother trend. It is observed that $\lambda_1$ and $\lambda_2$ in trend filtering are relatively insensitive. For the total number of iteration (the number to repeat steps (3)-(5)), 3 iterations are often enough for convergence.

\begin{figure}[t!]
    \centering
    \includegraphics[width=0.99\linewidth]{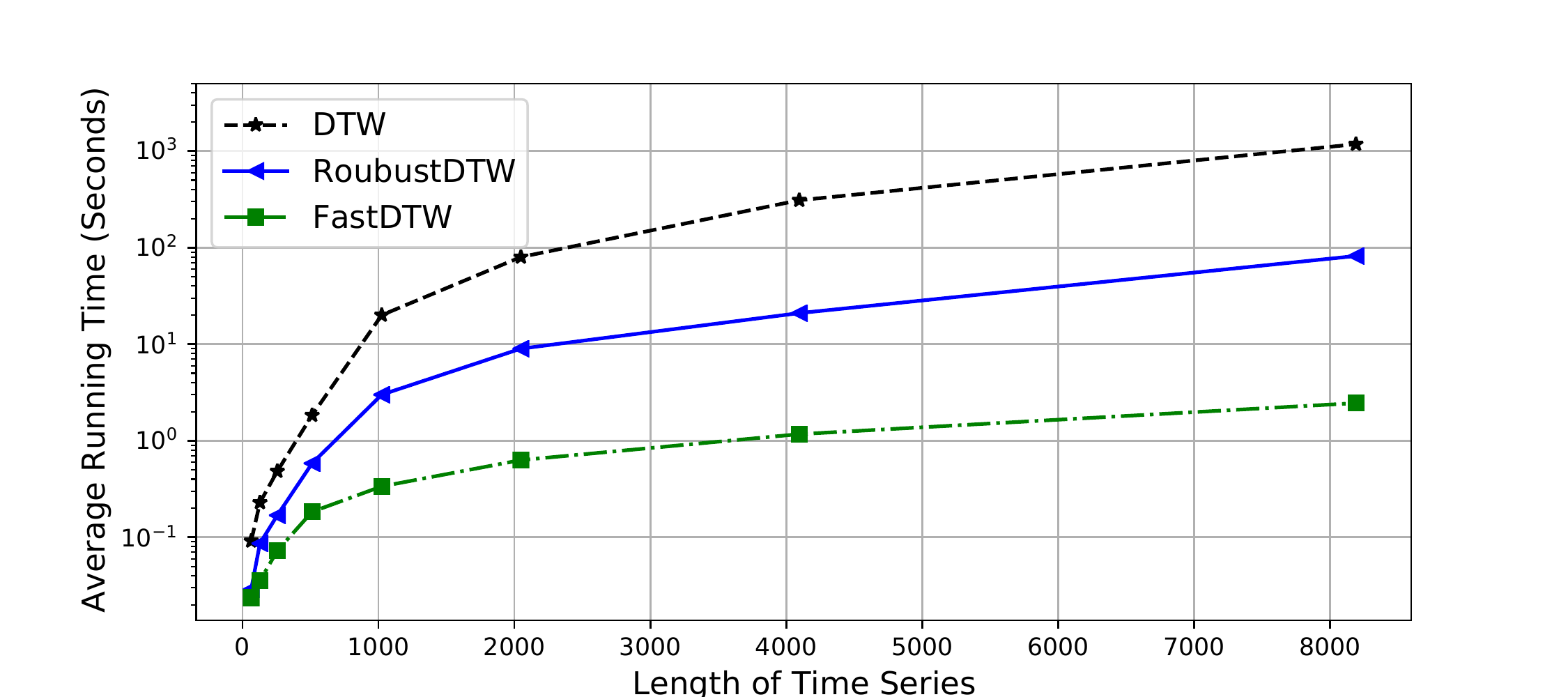}
    \vspace{-2mm}
    \caption{
    Running time (log scale) comparison of DTW, FastDTW, and RobustDTW under different lengths of time series.}
    \vspace{-4mm}
    \label{fig:5-run-time}
\end{figure}

To evaluate the efficiency, we summarize the average running time in log scale of DTW, FastDTW, and proposed RobustDTW in Fig.~\ref{fig:5-run-time} with synthetic datasets. The synthetic TS are single period of a sine wave with random noise and length varies from $2^6$ to $2^{13}$. The experiment is performed on a PC with 2.3GHz Intel Core i5 CPU and 16GB RAM. 
It can be observed in Fig.~\ref{fig:5-run-time} that the computational time of the standard DTW increases rapidly as the length of TS increases. Due to the proposed multi-level framework, our RobustDTW is significantly efficient than the standard DTW, especially for long time series. Although RobustDTW is slower than FastDTW, the speed of RobustDTW is reasonable for most time series application and we will demonstrate later that the proposed RobustDTW is more robust than FastDTW and exhibits much better performances in both outlier time series detection and periodicity detection.

\subsection{Outlier Time Series Detection} 

The purpose of outlier TS detection is to find one or more abnormal TS compare to other peers~\cite{blazquez2021review}. 
In this experiment, we test the the popular local outlier factor (LOF) algorithm~\cite{breunig2000lof,wang2011efficient,liu2020data}, which finds outliers by measuring the local deviation of a given data point with respect to its neighbors, with different distance measures for outlier time series detection.
We collected two real-world multivariate TS datasets from a top cloud computing company which are the measurement of the API response time (RT) and network speed (NetSpd) of two clusters, respectively. Specifically, The dataset ``RT" contains 294 TS and 1 of them are outliers; and the ``NetSpd" dataset contains 486 TS and 6 of them are outliers. We apply the LocalOutlierFactor function~\cite{LocalOutlierFactor} from scikit learn package to conduct LOF step for anomaly detection. For all different distance measures the hyper-parameters of LOF are tuned to get best performance, where the number of neighbors is set to be $30$ and contamination is $0.02$.

Table~\ref{tab:to_lof} compares the AUC scores of LOF algorithm with different distance measures, including Euclidean Distance (ED), standard DTW, FastDTW, and our RobustDTW, under different noise conditions, i.e, raw datasets, datasets with injected dips, datasets with injected spike, and datasets with injected both spike and dips. The standard DTW performs the second best, while FastDTW achieves fast computation at the cost of worse accuracy. In contrast, our RobustDTW achieves the best performance consistently.

\begin{table}[t!]
\centering
\caption{Comparison of AUC scores of outlier detection via LOF with different distance measures and noise conditions.}
\label{tab:to_lof}
\vspace{-3mm}
\scalebox{0.8}{
\begin{tabular}{|c|r|cccc|}
\hline
\small{Dataset}    & \small{Noise level}   & \small{ED}    & \small{DTW}   & \small{FastDTW} & \small\textbf{RobustDTW}     \\ \hline\hline
\multirow{4}{*}{RT} & \small{raw data}      & 0.887 & 0.986 & 0.949   & {\bf 0.997} \\ \cline{2-6} 
                   & \small{+ dips}        & 0.811 & 0.965 & 0.918   & {\bf 0.996} \\ \cline{2-6}
                   & \small{+ spikes}      & 0.362 & 0.939 & 0.775   & {\bf 0.972} \\ \cline{2-6} 
                   & \small{+ spikes \& dips}     & 0.317 & 0.580 & 0.655   & {\bf 0.969} \\ \hline\hline
\multirow{4}{*}{NetSpd} & \small{raw data}      & 0.674 & 0.982 & 0.917   & {\bf 0.988} \\ \cline{2-6} 
                   & \small{+ dips}        & 0.635 & 0.849 & 0.649   & {\bf 0.982} \\ \cline{2-6}
                   & \small{+ spikes}      & 0.642 & 0.915 & 0.874   & {\bf 0.975} \\ \cline{2-6} 
                   & \small{+ spikes \& dips}     & 0.508 & 0.627 & 0.616   & {\bf 0.938} \\ \hline
\end{tabular}
}
\vspace{-3mm}
\end{table}



\vspace{-1mm}
\subsection{Time Series Periodicity Detection} 
\begin{figure}[t!]
    \centering
    \includegraphics[width=0.99\linewidth]{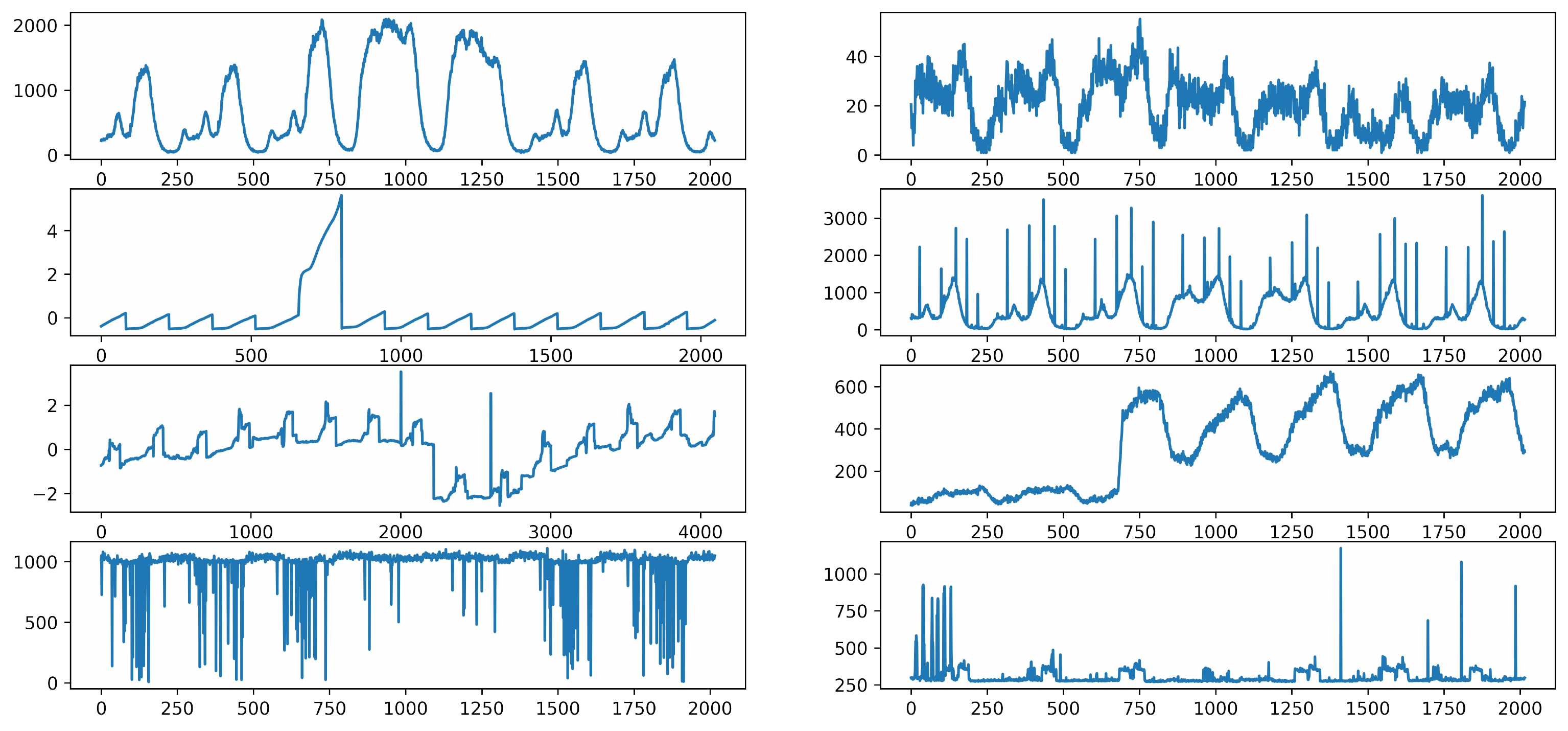}
    \vspace{-3mm}
    \caption{
    Real-world TS examples from cloud monitoring with complex patterns for periodicity detection. The TS in top three rows are periodic while the bottom ones are without periodicity. These time series have challenging noise, outliers, and/or trend changes.}
    \vspace{-1.5mm}
    \label{fig:4-pd-example}
\end{figure}

Many TS are characterized by repeating cycles, or periodicity. Periodicity detection aims at discovering the repeating patterns in TS. 
The periodicity detection is a nontrivial task~\cite{WenRobustPeriod20,vlachos2005periodicity} given complicated TS, e.g., non-stationary, sudden trend change, noise and outliers, as shown in Fig.~\ref{fig:4-pd-example}. 
Here we focus on dominant periodicity detection, where domain experts generally have the prior knowledge about the period length ($T$) if the TS are periodic.


\vspace{1mm}
{\hspace{-3.5mm}\bf RobustDTW-based Slicing Algorithm:}\\
In this part we propose a slicing algorithm to determine whether a given TS has a prior period of $T$ or not. This algorithm first detrends and normalizes input TS data, removes outliers, and then slices the TS into multiple segments with equal length of $T$. Next we calculate distance between adjacent segments using RobustDTW. 
For detrending, we apply the robust trend filtering~\cite{wen2019robusttrend} to remove the global trend which is useful for non-stationary TS.
For normalization, all TS are normalized by subtracting median and then dividing by the biweight scale. Note that median and biweight scale are more robust to noise and outliers compared to mean and standard deviation. The biweight scale~\cite{beers1990measures} is calculated as
\vspace{-1mm}
\begin{equation}\notag
\resizebox{0.9\hsize}{!}{$\xi_{biscl} \!=\! {\sqrt{n\sum_{|u_i|<1}(x_i-M)^2(1-u_i^2)^4}} \bigg/ {{\sqrt{\sum_{|u_i|<1}(1-u_i^2)(1-5u_i^2)}}}$},\vspace{-1mm}
\end{equation}
where $x$ is the input data, $M$ is the sample median, $u_i = (x_i - M)/(c*MAD)$ with
$c$ as the tuning constant typically set to be 9.0, and MAD is the median absolute deviation. 
After that, to further reduce the influence of outliers (such as the segment deviation due to black Friday), we filter out the outlier distances as determined by the ``$1.5\times$IQR rule'' (interquartile range).
We then calculate the mean value of the remaining distances which is sum of squared differences between values that aligned on both time series, as a measure of periodicity. And a TS is considered has a period of $T$ if the estimated mean value is less than a user-specified threshold.

\vspace{1mm}
{\hspace{-3.5mm}\bf Periodicity Detection Comparison:}\\
We collect 200 service monitoring time series from a top cloud computing company. Half of these TS are with daily periodicity. These TS data exhibit various shapes with challenging noise, outliers, and trend changes as shown in Fig.~\ref{fig:4-pd-example}. 
For model evaluation, we test our algorithm under different distance measures, including Euclidean distance, standard DTW, FastDTW and our RobustDTW. We also compare our method with some commonly used periodicity detection algorihtms such as ACF, as well as the state-of-the-art AUTOPERIOD~\cite{vlachos2005periodicity,Mitsa2010} and RobustPeriod~\cite{WenRobustPeriod20} algorithms.
The performance metrics including precision, recall and F1, are summarized in Table~\ref{tab:F1 of DTWs}, and the best performance is highlighted in bold. It can be observed that among all compared methods, our method using slicing and RobustDTW achieves the best F1 score. 
We attribute the advantages of our periodicity detection using RobustDTW to: 1) graph detrending can reduce the pairwise distance significantly when two TS have similar shape; 2) filtering neighboring distance can exclude large irregular shapes within one period; 3) robust normalization using median and biweight scale is insensitive to outliers and help parameter tuning easier.

\begin{table}[t!]
\centering
\caption{Comparison of periodicity detection methods.}\label{tab:F1 of DTWs}
\vspace{-3mm}
\scalebox{0.8}{
\begin{tabular}{|c|c|c|c|c|}
\hline
\multicolumn{2}{|c|}{Methods}     & Precision & Recall & F1 \\\hline \hline
\multicolumn{2}{|c|}{ACF}           & 0.960 & 0.701  & 0.810  \\ \hline 
\multicolumn{2}{|c|}{AUTOPERIOD~\cite{vlachos2005periodicity}}    & \textbf{0.980} & 0.715  & 0.827  \\ \hline 
\multicolumn{2}{|c|}{RobustPeriod~\cite{WenRobustPeriod20}}  & 0.920 & 0.902  & 0.911  \\ \hline 
\multirow{3}{*}{Slicing} & ED     & 0.919 &	0.800  & 0.855 \\ \cline{2-5} 
                                &  DTW  & 0.911 &	0.930  &	0.920 \\  \cline{2-5} 
                                & FastDTW      & 0.873 &	0.970  &	0.919 \\ \cline{2-5}
                       & \textbf{RobustDTW}    & 0.951 &  \textbf{0.980}  & \textbf{0.966}  \\ \hline 
\end{tabular}
}
\vspace{-4mm}
\end{table}

\section{Conclusion}

In this paper, we propose a novel RobustDTW measure which estimates the time warping function and the trend simultaneously by an alternating approach and we further accelerate it in a multi-level framework. Compared with existing DTW and its variants, it is fast and robust to noise and outliers. We demonstrate the successful applications of our RobustDTW algorithm in outlier time series detection and periodicity detection.



\bibliographystyle{ACM-Reference-Format}
\balance
\bibliography{6-reference}


\begin{thebibliography}{42}


\ifx \showCODEN    \undefined \def \showCODEN     #1{\unskip}     \fi
\ifx \showDOI      \undefined \def \showDOI       #1{#1}\fi
\ifx \showISBNx    \undefined \def \showISBNx     #1{\unskip}     \fi
\ifx \showISBNxiii \undefined \def \showISBNxiii  #1{\unskip}     \fi
\ifx \showISSN     \undefined \def \showISSN      #1{\unskip}     \fi
\ifx \showLCCN     \undefined \def \showLCCN      #1{\unskip}     \fi
\ifx \shownote     \undefined \def \shownote      #1{#1}          \fi
\ifx \showarticletitle \undefined \def \showarticletitle #1{#1}   \fi
\ifx \showURL      \undefined \def \showURL       {\relax}        \fi
\providecommand\bibfield[2]{#2}
\providecommand\bibinfo[2]{#2}
\providecommand\natexlab[1]{#1}
\providecommand\showeprint[2][]{arXiv:#2}

\bibitem[\protect\citeauthoryear{Alexandrov, Bianconcini, Dagum, Maass, and
  McElroy}{Alexandrov et~al\mbox{.}}{2012}]%
        {alexandrov2012review}
\bibfield{author}{\bibinfo{person}{Theodore Alexandrov},
  \bibinfo{person}{Silvia Bianconcini}, \bibinfo{person}{Estela~Bee Dagum},
  \bibinfo{person}{Peter Maass}, {and} \bibinfo{person}{Tucker~S McElroy}.}
  \bibinfo{year}{2012}\natexlab{}.
\newblock \showarticletitle{A review of some modern approaches to the problem
  of trend extraction}.
\newblock \bibinfo{journal}{\emph{Econometric Reviews}} \bibinfo{volume}{31},
  \bibinfo{number}{6} (\bibinfo{year}{2012}), \bibinfo{pages}{593--624}.
\newblock


\bibitem[\protect\citeauthoryear{Angelosante and Giannakis}{Angelosante and
  Giannakis}{2011}]%
        {angelosante2011sparse}
\bibfield{author}{\bibinfo{person}{Daniele Angelosante} {and}
  \bibinfo{person}{Georgios~B Giannakis}.} \bibinfo{year}{2011}\natexlab{}.
\newblock \showarticletitle{Sparse graphical modeling of piecewise-stationary
  time series}. In \bibinfo{booktitle}{\emph{2011 IEEE International Conference
  on Acoustics, Speech and Signal Processing (ICASSP)}}.
  \bibinfo{pages}{1960--1963}.
\newblock


\bibitem[\protect\citeauthoryear{Barbon, Guido, Vieira, {Silva Fonseca},
  Sanchez, Scalassara, Maciel, Pereira, and Chen}{Barbon et~al\mbox{.}}{2009}]%
        {wavelet:DTW}
\bibfield{author}{\bibinfo{person}{Sylvio Barbon},
  \bibinfo{person}{Rodrigo~Capobianco Guido}, \bibinfo{person}{Lucimar~Sasso
  Vieira}, \bibinfo{person}{Everthon {Silva Fonseca}},
  \bibinfo{person}{Fabrício~Lopes Sanchez}, \bibinfo{person}{Paulo~Rogério
  Scalassara}, \bibinfo{person}{Carlos~Dias Maciel},
  \bibinfo{person}{José~Carlos Pereira}, {and} \bibinfo{person}{Shi-Huang
  Chen}.} \bibinfo{year}{2009}\natexlab{}.
\newblock \showarticletitle{Wavelet-based dynamic time warping}.
\newblock \bibinfo{journal}{\emph{J. Comput. Appl. Math.}}
  \bibinfo{volume}{227}, \bibinfo{number}{2} (\bibinfo{year}{2009}),
  \bibinfo{pages}{271 -- 287}.
\newblock
\newblock
\shownote{Special Issue on Emergent Applications of Fractals and Wavelets in
  Biology and Biomedicine.}


\bibitem[\protect\citeauthoryear{Beers, Flynn, and Gebhardt}{Beers
  et~al\mbox{.}}{1990}]%
        {beers1990measures}
\bibfield{author}{\bibinfo{person}{Timothy~C Beers}, \bibinfo{person}{Kevin
  Flynn}, {and} \bibinfo{person}{Karl Gebhardt}.}
  \bibinfo{year}{1990}\natexlab{}.
\newblock \showarticletitle{Measures of location and scale for velocities in
  clusters of galaxies-A robust approach}.
\newblock \bibinfo{journal}{\emph{The Astronomical Journal}}
  \bibinfo{volume}{100} (\bibinfo{year}{1990}), \bibinfo{pages}{32--46}.
\newblock


\bibitem[\protect\citeauthoryear{Benkabou, Benabdeslem, and Canitia}{Benkabou
  et~al\mbox{.}}{2018}]%
        {benkabou2018unsupervised}
\bibfield{author}{\bibinfo{person}{Seif-Eddine Benkabou},
  \bibinfo{person}{Khalid Benabdeslem}, {and} \bibinfo{person}{Bruno Canitia}.}
  \bibinfo{year}{2018}\natexlab{}.
\newblock \showarticletitle{Unsupervised outlier detection for time series by
  entropy and dynamic time warping}.
\newblock \bibinfo{journal}{\emph{Knowledge and Information Systems}}
  \bibinfo{volume}{54}, \bibinfo{number}{2} (\bibinfo{year}{2018}),
  \bibinfo{pages}{463--486}.
\newblock


\bibitem[\protect\citeauthoryear{Bl{\'a}zquez-Garc{\'\i}a, Conde, Mori, and
  Lozano}{Bl{\'a}zquez-Garc{\'\i}a et~al\mbox{.}}{2021}]%
        {blazquez2021review}
\bibfield{author}{\bibinfo{person}{Ane Bl{\'a}zquez-Garc{\'\i}a},
  \bibinfo{person}{Angel Conde}, \bibinfo{person}{Usue Mori}, {and}
  \bibinfo{person}{Jose~A Lozano}.} \bibinfo{year}{2021}\natexlab{}.
\newblock \showarticletitle{A Review on outlier/Anomaly Detection in Time
  Series Data}.
\newblock \bibinfo{journal}{\emph{ACM Computing Surveys (CSUR)}}
  \bibinfo{volume}{54}, \bibinfo{number}{3} (\bibinfo{year}{2021}),
  \bibinfo{pages}{1--33}.
\newblock


\bibitem[\protect\citeauthoryear{Boyd, Parikh, Chu, Peleato, and Eckstein}{Boyd
  et~al\mbox{.}}{2011}]%
        {boyd2011distributed}
\bibfield{author}{\bibinfo{person}{Stephen Boyd}, \bibinfo{person}{Neal
  Parikh}, \bibinfo{person}{Eric Chu}, \bibinfo{person}{Borja Peleato}, {and}
  \bibinfo{person}{Jonathan Eckstein}.} \bibinfo{year}{2011}\natexlab{}.
\newblock \showarticletitle{Distributed Optimization and Statistical Learning
  via the Alternating Direction Method of Multipliers}.
\newblock \bibinfo{journal}{\emph{Foundations and Trends{\textregistered} in
  Machine Learning}} \bibinfo{volume}{3}, \bibinfo{number}{1}
  (\bibinfo{year}{2011}), \bibinfo{pages}{1--122}.
\newblock


\bibitem[\protect\citeauthoryear{Breunig, Kriegel, Ng, and Sander}{Breunig
  et~al\mbox{.}}{2000}]%
        {breunig2000lof}
\bibfield{author}{\bibinfo{person}{Markus~M Breunig},
  \bibinfo{person}{Hans-Peter Kriegel}, \bibinfo{person}{Raymond~T Ng}, {and}
  \bibinfo{person}{J{\"o}rg Sander}.} \bibinfo{year}{2000}\natexlab{}.
\newblock \showarticletitle{LOF: identifying density-based local outliers}. In
  \bibinfo{booktitle}{\emph{Proceedings of the 2000 ACM SIGMOD international
  conference on Management of data}}. \bibinfo{pages}{93--104}.
\newblock


\bibitem[\protect\citeauthoryear{Chang, Tung, and Mori}{Chang
  et~al\mbox{.}}{2021}]%
        {chang2021learning}
\bibfield{author}{\bibinfo{person}{Xiaobin Chang}, \bibinfo{person}{Frederick
  Tung}, {and} \bibinfo{person}{Greg Mori}.} \bibinfo{year}{2021}\natexlab{}.
\newblock \showarticletitle{Learning discriminative prototypes with dynamic
  time warping}. In \bibinfo{booktitle}{\emph{Proceedings of the IEEE/CVF
  Conference on Computer Vision and Pattern Recognition (CVPR)}}.
  \bibinfo{pages}{8395--8404}.
\newblock


\bibitem[\protect\citeauthoryear{Cuturi and Blondel}{Cuturi and
  Blondel}{2017}]%
        {cuturi2017soft}
\bibfield{author}{\bibinfo{person}{Marco Cuturi} {and} \bibinfo{person}{Mathieu
  Blondel}.} \bibinfo{year}{2017}\natexlab{}.
\newblock \showarticletitle{{Soft-DTW}: a differentiable loss function for
  time-series}. In \bibinfo{booktitle}{\emph{International conference on
  machine learning (ICML)}}. \bibinfo{pages}{894--903}.
\newblock


\bibitem[\protect\citeauthoryear{Deriso and Boyd}{Deriso and Boyd}{2019}]%
        {Boyd:DTW:2020}
\bibfield{author}{\bibinfo{person}{Dave Deriso} {and}
  \bibinfo{person}{Stephen~P. Boyd}.} \bibinfo{year}{2019}\natexlab{}.
\newblock \showarticletitle{A General Optimization Framework for Dynamic Time
  Warping}.
\newblock \bibinfo{journal}{\emph{CoRR}}  \bibinfo{volume}{abs/1905.12893}
  (\bibinfo{year}{2019}).
\newblock
\showeprint[arxiv]{1905.12893}


\bibitem[\protect\citeauthoryear{{Diab}, {AsSadhan}, {Binsalleeh},
  {Lambotharan}, {Kyriakopoulos}, and {Ghafir}}{{Diab} et~al\mbox{.}}{2019}]%
        {DTW:outlier:2019:Diab}
\bibfield{author}{\bibinfo{person}{D.~M. {Diab}}, \bibinfo{person}{B.
  {AsSadhan}}, \bibinfo{person}{H. {Binsalleeh}}, \bibinfo{person}{S.
  {Lambotharan}}, \bibinfo{person}{K.~G. {Kyriakopoulos}}, {and}
  \bibinfo{person}{I. {Ghafir}}.} \bibinfo{year}{2019}\natexlab{}.
\newblock \showarticletitle{Anomaly Detection Using Dynamic Time Warping}. In
  \bibinfo{booktitle}{\emph{2019 IEEE International Conference on Computational
  Science and Engineering (CSE) and IEEE International Conference on Embedded
  and Ubiquitous Computing (EUC)}}. \bibinfo{pages}{193--198}.
\newblock


\bibitem[\protect\citeauthoryear{Elfeky, Aref, and Elmagarmid}{Elfeky
  et~al\mbox{.}}{2005}]%
        {elfeky2005warp}
\bibfield{author}{\bibinfo{person}{Mohamed~G Elfeky}, \bibinfo{person}{Walid~G
  Aref}, {and} \bibinfo{person}{Ahmed~K Elmagarmid}.}
  \bibinfo{year}{2005}\natexlab{}.
\newblock \showarticletitle{WARP: time warping for periodicity detection}. In
  \bibinfo{booktitle}{\emph{Fifth IEEE International Conference on Data Mining
  (ICDM'05)}}. \bibinfo{pages}{8--pp}.
\newblock


\bibitem[\protect\citeauthoryear{Esling and Agon}{Esling and Agon}{2012}]%
        {esling2012time}
\bibfield{author}{\bibinfo{person}{Philippe Esling} {and}
  \bibinfo{person}{Carlos Agon}.} \bibinfo{year}{2012}\natexlab{}.
\newblock \showarticletitle{Time-series data mining}.
\newblock \bibinfo{journal}{\emph{ACM Computing Surveys (CSUR)}}
  \bibinfo{volume}{45}, \bibinfo{number}{1} (\bibinfo{year}{2012}),
  \bibinfo{pages}{1--34}.
\newblock


\bibitem[\protect\citeauthoryear{Han, Li, Gao, and Wang}{Han
  et~al\mbox{.}}{2018}]%
        {han2018accurate}
\bibfield{author}{\bibinfo{person}{Renmin Han}, \bibinfo{person}{Yu Li},
  \bibinfo{person}{Xin Gao}, {and} \bibinfo{person}{Sheng Wang}.}
  \bibinfo{year}{2018}\natexlab{}.
\newblock \showarticletitle{An accurate and rapid continuous wavelet dynamic
  time warping algorithm for end-to-end mapping in ultra-long nanopore
  sequencing}.
\newblock \bibinfo{journal}{\emph{Bioinformatics}} \bibinfo{volume}{34},
  \bibinfo{number}{17} (\bibinfo{year}{2018}), \bibinfo{pages}{i722--i731}.
\newblock


\bibitem[\protect\citeauthoryear{Hodrick and Prescott}{Hodrick and
  Prescott}{1997}]%
        {hodrick1997postwar}
\bibfield{author}{\bibinfo{person}{Robert~J Hodrick} {and}
  \bibinfo{person}{Edward~C Prescott}.} \bibinfo{year}{1997}\natexlab{}.
\newblock \showarticletitle{Postwar US business cycles: an empirical
  investigation}.
\newblock \bibinfo{journal}{\emph{Journal of Money, credit, and Banking}}
  (\bibinfo{year}{1997}), \bibinfo{pages}{1--16}.
\newblock


\bibitem[\protect\citeauthoryear{Huang, Wang, Wu, and Tang}{Huang
  et~al\mbox{.}}{2019}]%
        {huang2019dsanet}
\bibfield{author}{\bibinfo{person}{Siteng Huang}, \bibinfo{person}{Donglin
  Wang}, \bibinfo{person}{Xuehan Wu}, {and} \bibinfo{person}{Ao Tang}.}
  \bibinfo{year}{2019}\natexlab{}.
\newblock \showarticletitle{Dsanet: Dual self-attention network for
  multivariate time series forecasting}. In
  \bibinfo{booktitle}{\emph{Proceedings of the 28th ACM International
  Conference on Information and Knowledge Management (CIKM)}}.
  \bibinfo{pages}{2129--2132}.
\newblock


\bibitem[\protect\citeauthoryear{Jeong, Jeong, and Omitaomu}{Jeong
  et~al\mbox{.}}{2011}]%
        {WDTW:classification:PR:2011}
\bibfield{author}{\bibinfo{person}{Young-Seon Jeong}, \bibinfo{person}{Myong~K.
  Jeong}, {and} \bibinfo{person}{Olufemi~A. Omitaomu}.}
  \bibinfo{year}{2011}\natexlab{}.
\newblock \showarticletitle{Weighted dynamic time warping for time series
  classification}.
\newblock \bibinfo{journal}{\emph{Pattern Recognition}} \bibinfo{volume}{44},
  \bibinfo{number}{9} (\bibinfo{year}{2011}), \bibinfo{pages}{2231 -- 2240}.
\newblock
\newblock
\shownote{Computer Analysis of Images and Patterns.}


\bibitem[\protect\citeauthoryear{Junior, Guido, Chen, Vieira, and
  Sanchez}{Junior et~al\mbox{.}}{2007}]%
        {junior2007improved}
\bibfield{author}{\bibinfo{person}{Sylvio~Barbon Junior},
  \bibinfo{person}{Rodrigo~Capobianco Guido}, \bibinfo{person}{Shi-Huang Chen},
  \bibinfo{person}{Lucimar~Sasso Vieira}, {and} \bibinfo{person}{Fabricio~Lopes
  Sanchez}.} \bibinfo{year}{2007}\natexlab{}.
\newblock \showarticletitle{Improved dynamic time warping based on the discrete
  wavelet transform}. In \bibinfo{booktitle}{\emph{Ninth IEEE International
  Symposium on Multimedia Workshops (ISMW 2007)}}. \bibinfo{pages}{256--263}.
\newblock


\bibitem[\protect\citeauthoryear{Keogh and Ratanamahatana}{Keogh and
  Ratanamahatana}{2005}]%
        {keogh2005exact}
\bibfield{author}{\bibinfo{person}{Eamonn Keogh} {and}
  \bibinfo{person}{Chotirat~Ann Ratanamahatana}.}
  \bibinfo{year}{2005}\natexlab{}.
\newblock \showarticletitle{Exact indexing of dynamic time warping}.
\newblock \bibinfo{journal}{\emph{Knowledge and information systems}}
  \bibinfo{volume}{7}, \bibinfo{number}{3} (\bibinfo{year}{2005}),
  \bibinfo{pages}{358--386}.
\newblock


\bibitem[\protect\citeauthoryear{Keogh and Pazzani}{Keogh and Pazzani}{2001}]%
        {keogh2001derivative}
\bibfield{author}{\bibinfo{person}{Eamonn~J Keogh} {and}
  \bibinfo{person}{Michael~J Pazzani}.} \bibinfo{year}{2001}\natexlab{}.
\newblock \showarticletitle{Derivative dynamic time warping}. In
  \bibinfo{booktitle}{\emph{Proceedings of the 2001 SIAM international
  conference on data mining}}. \bibinfo{pages}{1--11}.
\newblock


\bibitem[\protect\citeauthoryear{Kim, Koh, Boyd, and Gorinevsky}{Kim
  et~al\mbox{.}}{2009}]%
        {kim2009ell_1}
\bibfield{author}{\bibinfo{person}{Seung-Jean Kim}, \bibinfo{person}{Kwangmoo
  Koh}, \bibinfo{person}{Stephen Boyd}, {and} \bibinfo{person}{Dimitry
  Gorinevsky}.} \bibinfo{year}{2009}\natexlab{}.
\newblock \showarticletitle{$ell\_1$ trend filtering}.
\newblock \bibinfo{journal}{\emph{SIAM review}} \bibinfo{volume}{51},
  \bibinfo{number}{2} (\bibinfo{year}{2009}), \bibinfo{pages}{339--360}.
\newblock


\bibitem[\protect\citeauthoryear{Li, Liu, Yang, Liu, Wu, and Wan}{Li
  et~al\mbox{.}}{2020}]%
        {LI2020:adaptive:dtw}
\bibfield{author}{\bibinfo{person}{Huanhuan Li}, \bibinfo{person}{Jingxian
  Liu}, \bibinfo{person}{Zaili Yang}, \bibinfo{person}{Ryan~Wen Liu},
  \bibinfo{person}{Kefeng Wu}, {and} \bibinfo{person}{Yuan Wan}.}
  \bibinfo{year}{2020}\natexlab{}.
\newblock \showarticletitle{Adaptively constrained dynamic time warping for
  time series classification and clustering}.
\newblock \bibinfo{journal}{\emph{Information Sciences}}  \bibinfo{volume}{534}
  (\bibinfo{year}{2020}), \bibinfo{pages}{97 -- 116}.
\newblock


\bibitem[\protect\citeauthoryear{Liu, Zhao, Lin, Liu, Ding, Yang, and Yi}{Liu
  et~al\mbox{.}}{2020}]%
        {liu2020data}
\bibfield{author}{\bibinfo{person}{Shengyuan Liu}, \bibinfo{person}{Yuxuan
  Zhao}, \bibinfo{person}{Zhenzhi Lin}, \bibinfo{person}{Yilu Liu},
  \bibinfo{person}{Yi Ding}, \bibinfo{person}{Li Yang}, {and}
  \bibinfo{person}{Shimin Yi}.} \bibinfo{year}{2020}\natexlab{}.
\newblock \showarticletitle{Data-driven event detection of power systems based
  on unequal-interval reduction of PMU data and local outlier factor}.
\newblock \bibinfo{journal}{\emph{IEEE Transactions on Smart Grid}}
  \bibinfo{volume}{11}, \bibinfo{number}{2} (\bibinfo{year}{2020}),
  \bibinfo{pages}{1630--1643}.
\newblock


\bibitem[\protect\citeauthoryear{{LocalOutlierFactor}}{{LocalOutlierFactor}}{2021}]%
        {LocalOutlierFactor}
\bibfield{author}{\bibinfo{person}{{LocalOutlierFactor}}.}
  \bibinfo{year}{2021}\natexlab{}.
\newblock
  \bibinfo{howpublished}{\url{https://scikit-learn.org/stable/modules/generated/sklearn.neighbors.LocalOutlierFactor.html}}.
\newblock
\newblock
\shownote{[Online; accessed May-2021].}


\bibitem[\protect\citeauthoryear{Mitsa}{Mitsa}{2010}]%
        {Mitsa2010}
\bibfield{author}{\bibinfo{person}{Theophano Mitsa}.}
  \bibinfo{year}{2010}\natexlab{}.
\newblock \bibinfo{booktitle}{\emph{{Temporal Data Mining}}
  (\bibinfo{edition}{1st} ed.)}.
\newblock \bibinfo{publisher}{Chapman and Hall/CRC}, \bibinfo{address}{Boca
  Raton, FL}.
\newblock
\showISBNx{1420089765, 9781420089769}


\bibitem[\protect\citeauthoryear{Paparrizos, Liu, Elmore, and
  Franklin}{Paparrizos et~al\mbox{.}}{2020}]%
        {paparrizos2020debunking}
\bibfield{author}{\bibinfo{person}{John Paparrizos}, \bibinfo{person}{Chunwei
  Liu}, \bibinfo{person}{Aaron~J Elmore}, {and} \bibinfo{person}{Michael~J
  Franklin}.} \bibinfo{year}{2020}\natexlab{}.
\newblock \showarticletitle{Debunking four long-standing misconceptions of
  time-series distance measures}. In \bibinfo{booktitle}{\emph{Proceedings of
  the 2020 ACM SIGMOD International Conference on Management of Data}}.
  \bibinfo{pages}{1887--1905}.
\newblock


\bibitem[\protect\citeauthoryear{Pr{\"a}tzlich, Driedger, and
  M{\"u}ller}{Pr{\"a}tzlich et~al\mbox{.}}{2016}]%
        {pratzlich2016memory}
\bibfield{author}{\bibinfo{person}{Thomas Pr{\"a}tzlich},
  \bibinfo{person}{Jonathan Driedger}, {and} \bibinfo{person}{Meinard
  M{\"u}ller}.} \bibinfo{year}{2016}\natexlab{}.
\newblock \showarticletitle{Memory-restricted multiscale dynamic time warping}.
  In \bibinfo{booktitle}{\emph{2016 IEEE International Conference on Acoustics,
  Speech and Signal Processing (ICASSP)}}. \bibinfo{pages}{569--573}.
\newblock


\bibitem[\protect\citeauthoryear{Rakthanmanon, Campana, Mueen, Batista,
  Westover, Zhu, Zakaria, and Keogh}{Rakthanmanon et~al\mbox{.}}{2012}]%
        {search:dtw:kdd:2012}
\bibfield{author}{\bibinfo{person}{Thanawin Rakthanmanon},
  \bibinfo{person}{Bilson Campana}, \bibinfo{person}{Abdullah Mueen},
  \bibinfo{person}{Gustavo Batista}, \bibinfo{person}{Brandon Westover},
  \bibinfo{person}{Qiang Zhu}, \bibinfo{person}{Jesin Zakaria}, {and}
  \bibinfo{person}{Eamonn Keogh}.} \bibinfo{year}{2012}\natexlab{}.
\newblock \showarticletitle{Searching and Mining Trillions of Time Series
  Subsequences under Dynamic Time Warping}. In
  \bibinfo{booktitle}{\emph{Proceedings of the 18th ACM SIGKDD International
  Conference on Knowledge Discovery and Data Mining}}
  \emph{(\bibinfo{series}{KDD '12})}. \bibinfo{pages}{262–270}.
\newblock


\bibitem[\protect\citeauthoryear{Sakoe and Chiba}{Sakoe and Chiba}{1971}]%
        {SakoeChiba71}
\bibfield{author}{\bibinfo{person}{Hiroaki Sakoe} {and} \bibinfo{person}{Seibi
  Chiba}.} \bibinfo{year}{1971}\natexlab{}.
\newblock \showarticletitle{A Dynamic Programming Approach to Continuous Speech
  Recognition}. In \bibinfo{booktitle}{\emph{Proceedings of the Seventh
  International Congress on Acoustics, Budapest}}, Vol.~\bibinfo{volume}{3}.
  \bibinfo{publisher}{{Akad\'{e}miai} {Kiad\'{o}}},
  \bibinfo{address}{Budapest}, \bibinfo{pages}{65--69}.
\newblock


\bibitem[\protect\citeauthoryear{Salvador and Chan}{Salvador and Chan}{2007}]%
        {salvador2007toward}
\bibfield{author}{\bibinfo{person}{Stan Salvador} {and} \bibinfo{person}{Philip
  Chan}.} \bibinfo{year}{2007}\natexlab{}.
\newblock \showarticletitle{FastDTW: Toward accurate dynamic time warping in
  linear time and space}.
\newblock \bibinfo{journal}{\emph{Intelligent Data Analysis}}
  \bibinfo{volume}{11}, \bibinfo{number}{5} (\bibinfo{year}{2007}),
  \bibinfo{pages}{561--580}.
\newblock


\bibitem[\protect\citeauthoryear{Vlachos, Yu, and Castelli}{Vlachos
  et~al\mbox{.}}{2005}]%
        {vlachos2005periodicity}
\bibfield{author}{\bibinfo{person}{Michail Vlachos}, \bibinfo{person}{Philip
  Yu}, {and} \bibinfo{person}{Vittorio Castelli}.}
  \bibinfo{year}{2005}\natexlab{}.
\newblock \showarticletitle{On periodicity detection and structural periodic
  similarity}. In \bibinfo{booktitle}{\emph{Proceedings of the 2005 SIAM
  international conference on data mining}}. \bibinfo{pages}{449--460}.
\newblock


\bibitem[\protect\citeauthoryear{Wang and Lu}{Wang and Lu}{2011}]%
        {wang2011efficient}
\bibfield{author}{\bibinfo{person}{Wei Wang} {and} \bibinfo{person}{Peizhong
  Lu}.} \bibinfo{year}{2011}\natexlab{}.
\newblock \showarticletitle{An efficient switching median filter based on local
  outlier factor}.
\newblock \bibinfo{journal}{\emph{IEEE Signal Processing Letters}}
  \bibinfo{volume}{18}, \bibinfo{number}{10} (\bibinfo{year}{2011}),
  \bibinfo{pages}{551--554}.
\newblock


\bibitem[\protect\citeauthoryear{Wang, Sharpnack, Smola, and Tibshirani}{Wang
  et~al\mbox{.}}{2016}]%
        {wang2016trend}
\bibfield{author}{\bibinfo{person}{Yu-Xiang Wang}, \bibinfo{person}{James
  Sharpnack}, \bibinfo{person}{Alexander~J Smola}, {and}
  \bibinfo{person}{Ryan~J Tibshirani}.} \bibinfo{year}{2016}\natexlab{}.
\newblock \showarticletitle{Trend filtering on graphs}.
\newblock \bibinfo{journal}{\emph{The Journal of Machine Learning Research}}
  \bibinfo{volume}{17}, \bibinfo{number}{1} (\bibinfo{year}{2016}),
  \bibinfo{pages}{3651--3691}.
\newblock


\bibitem[\protect\citeauthoryear{Wen, Gao, Song, Sun, and Tan}{Wen
  et~al\mbox{.}}{2019}]%
        {wen2019robusttrend}
\bibfield{author}{\bibinfo{person}{Qingsong Wen}, \bibinfo{person}{Jingkun
  Gao}, \bibinfo{person}{Xiaomin Song}, \bibinfo{person}{Liang Sun}, {and}
  \bibinfo{person}{Jian Tan}.} \bibinfo{year}{2019}\natexlab{}.
\newblock \showarticletitle{{RobustTrend}: a {Huber} loss with a combined first
  and second order difference regularization for time series trend filtering}.
  In \bibinfo{booktitle}{\emph{Proceedings of the 28th International Joint
  Conference on Artificial Intelligence (IJCAI'19)}}. AAAI Press,
  \bibinfo{pages}{3856--3862}.
\newblock


\bibitem[\protect\citeauthoryear{Wen, He, Sun, Zhang, Ke, and Xu}{Wen
  et~al\mbox{.}}{2021}]%
        {WenRobustPeriod20}
\bibfield{author}{\bibinfo{person}{Qingsong Wen}, \bibinfo{person}{Kai He},
  \bibinfo{person}{Liang Sun}, \bibinfo{person}{Yingying Zhang},
  \bibinfo{person}{Min Ke}, {and} \bibinfo{person}{Huan Xu}.}
  \bibinfo{year}{2021}\natexlab{}.
\newblock \showarticletitle{{RobustPeriod}: Time-Frequency Mining for Robust
  Multiple Periodicity Detection}. In \bibinfo{booktitle}{\emph{Proceedings of
  ACM International Conference on Management of Data (SIGMOD'21)}}.
  \bibinfo{pages}{2328--2337}.
\newblock


\bibitem[\protect\citeauthoryear{Wen, Zhang, Li, and Sun}{Wen
  et~al\mbox{.}}{2020}]%
        {wen2020fastrobustSTL}
\bibfield{author}{\bibinfo{person}{Qingsong Wen}, \bibinfo{person}{Zhe Zhang},
  \bibinfo{person}{Yan Li}, {and} \bibinfo{person}{Liang Sun}.}
  \bibinfo{year}{2020}\natexlab{}.
\newblock \showarticletitle{Fast {RobustSTL}: Efficient and Robust
  Seasonal-Trend Decomposition for Time Series with Complex Patterns}. In
  \bibinfo{booktitle}{\emph{Proceedings of the 26th ACM SIGKDD International
  Conference on Knowledge Discovery \& Data Mining (KDD'20)}}.
  \bibinfo{pages}{2203--2213}.
\newblock


\bibitem[\protect\citeauthoryear{Wen and Zeng}{Wen and Zeng}{1999}]%
        {Wen:median:filter}
\bibfield{author}{\bibinfo{person}{Yi Wen} {and} \bibinfo{person}{Bing Zeng}.}
  \bibinfo{year}{1999}\natexlab{}.
\newblock \showarticletitle{A simple nonlinear filter for economic time series
  analysis}.
\newblock \bibinfo{journal}{\emph{Economics Letters}} \bibinfo{volume}{64},
  \bibinfo{number}{2} (\bibinfo{year}{1999}), \bibinfo{pages}{151--160}.
\newblock
\showISSN{0165-1765}
\urldef\tempurl%
\url{https://doi.org/10.1016/S0165-1765(99)00089-0}
\showDOI{\tempurl}


\bibitem[\protect\citeauthoryear{Wu and Keogh}{Wu and Keogh}{2020}]%
        {wu2020fastdtw}
\bibfield{author}{\bibinfo{person}{Renjie Wu} {and} \bibinfo{person}{Eamonn~J
  Keogh}.} \bibinfo{year}{2020}\natexlab{}.
\newblock \showarticletitle{FastDTW is approximate and Generally Slower than
  the Algorithm it Approximates}.
\newblock \bibinfo{journal}{\emph{arXiv preprint arXiv:2003.11246}}
  (\bibinfo{year}{2020}).
\newblock


\bibitem[\protect\citeauthoryear{Yuan, Lin, Zhang, and Wang}{Yuan
  et~al\mbox{.}}{2019}]%
        {yuan2019locally}
\bibfield{author}{\bibinfo{person}{Jidong Yuan}, \bibinfo{person}{Qianhong
  Lin}, \bibinfo{person}{Wei Zhang}, {and} \bibinfo{person}{Zhihai Wang}.}
  \bibinfo{year}{2019}\natexlab{}.
\newblock \showarticletitle{Locally slope-based dynamic time warping for time
  series classification}. In \bibinfo{booktitle}{\emph{Proceedings of the 28th
  ACM International Conference on Information and Knowledge Management
  (CIKM)}}. \bibinfo{pages}{1713--1722}.
\newblock


\bibitem[\protect\citeauthoryear{Zhang, Adl, and Glass}{Zhang
  et~al\mbox{.}}{2012}]%
        {zhang2012fast}
\bibfield{author}{\bibinfo{person}{Yaodong Zhang}, \bibinfo{person}{Kiarash
  Adl}, {and} \bibinfo{person}{James Glass}.} \bibinfo{year}{2012}\natexlab{}.
\newblock \showarticletitle{Fast spoken query detection using lower-bound
  dynamic time warping on graphical processing units}. In
  \bibinfo{booktitle}{\emph{2012 IEEE International Conference on Acoustics,
  Speech and Signal Processing (ICASSP)}}. \bibinfo{pages}{5173--5176}.
\newblock


\bibitem[\protect\citeauthoryear{Zhang, Guan, Qian, Xu, Liu, Wen, Sun, Jiang,
  Fan, and Ke}{Zhang et~al\mbox{.}}{2021}]%
        {zhang2021cloudrca}
\bibfield{author}{\bibinfo{person}{Yingying Zhang}, \bibinfo{person}{Zhengxiong
  Guan}, \bibinfo{person}{Huajie Qian}, \bibinfo{person}{Leili Xu},
  \bibinfo{person}{Hengbo Liu}, \bibinfo{person}{Qingsong Wen},
  \bibinfo{person}{Liang Sun}, \bibinfo{person}{Junwei Jiang},
  \bibinfo{person}{Lunting Fan}, {and} \bibinfo{person}{Min Ke}.}
  \bibinfo{year}{2021}\natexlab{}.
\newblock \showarticletitle{{CloudRCA}: A Root Cause Analysis Framework for
  Cloud Computing Platforms}. In \bibinfo{booktitle}{\emph{Proceedings of the
  30th ACM International Conference on Information \& Knowledge Management
  (CIKM'21)}}. \bibinfo{pages}{4373--4382}.
\newblock


\end{thebibliography}

\end{document}